\documentclass[12pt,a4paper]{report}

\usepackage{graphicx}
\usepackage{amssymb}
\usepackage{amsmath}
\usepackage{hyperref}
\usepackage{subcaption} 

\usepackage{framed}
\usepackage[margin=3.5cm]{geometry}

\usepackage[table,xcdraw]{xcolor}

\begin{document}

\begin{titlepage}
    \newgeometry{left=3cm,bottom=0.1cm,top=2.5cm}
    \begin{center}
        \textbf{ECOLE POLYTECHNIQUE FEDERALE DE LAUSANNE}
        \par
        \textbf{SCHOOL OF COMPUTER AND COMMUNICATION SCIENCES}

	\vspace{1cm}
	
        \begin{figure}[h]
        \centering
                \begin{minipage}{0.5\textwidth}
                \centering
                    \includegraphics[width=0.9\linewidth]{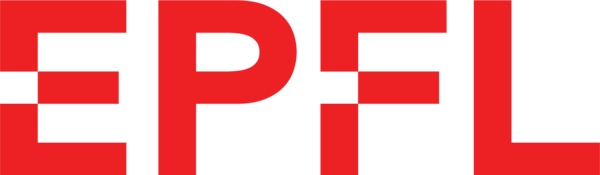}
                \end{minipage}%
                \begin{minipage}{0.5\textwidth}
                \centering
                    \includegraphics[width=0.9\linewidth]{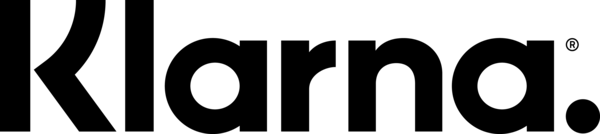}
                \end{minipage}
        \end{figure}

        Master project in Computer Science\par
        \vspace{1cm}
        {\scshape\Large\textbf{Learning Context-Aware Representations of Subtrees}\par}
        \vspace{2cm}
    
        Author\\
        \begin{large}
            \textbf{CEDRIC COOK}
        \end{large}

\vspace{8cm}
\begin{minipage}{0.5\textwidth}
Under the supervision of: \\
Dr. \textbf{Stefan Magureanu} \\
Klarna Bank AB
\end{minipage}%
\begin{minipage}{0.5\textwidth}
\begin{flushright}
Under the direction of: \\
Prof. \textbf{Boi Faltings} \\
Artificial Intelligence Laboratory, EPFL
\end{flushright}
\end{minipage}

    	\vspace{3cm}
        {\large Stockholm, March 22, 2019\par}
    \end{center}
\end{titlepage}
\restoregeometry

\setcounter{secnumdepth}{2}
\setcounter{tocdepth}{1}
\tableofcontents

\begin{abstract}
This thesis tackles the problem of learning efficient representations of complex, structured data with a natural application to web page and element classification. We hypothesise that the context around the element inside the web page is of high value to the problem and is currently under exploited. This thesis aims to solve the problem of classifying web elements as subtrees of a DOM tree by also considering their context.

To achieve this, first we discuss current expert knowledge systems that work on structures, such as Tree-LSTM. Then, we propose context-aware extensions to this model. We show that the new model achieves an average F1-score of 0.7973 on a multi-class web classification task. This model generates better representations for various subtrees and may be used for applications such element classification, state estimators in reinforcement learning over the Web and more.
\end{abstract}

\newpage

\section{Introduction} 
With the current omnipresence of the web, content is presented in more and more elaborate ways. Web pages are often segmented in visually or semantically distinguishable regions. Such regions may be navigational menus, content sections, interactable elements, advertisements, sitemaps and more. The principle content of a page may be further split up into individual elements. A social media website's content feed could be split up into individual posts, each of which may have a title, author name or link, timestamp, number of interactions and the content of the post. A web shop may have a product page which contains the name of the product, a photo of the product, size picker, a price and a button to add to cart for example.

In order to extract information from the web page, such as the content of the social media post or the price of the product, the element must first be properly identified. Many tree-based approaches for web page analysis and information extraction are attribute-based or depend on handcrafted heuristics. Such models often do not take into account all content and layout related information that are necessary to build a complete web page model. Furthermore, as such models are often built for a limited specific target set, they generalize poorly.

Another domain of high interest is reinforcement learning on the web. This may be used to learn to navigate workflows that are normally intended for humans, such as booking a flight or submitting an expense report. Specifically in the domain of reinforcement learning on the internet, estimating the state is difficult because of the input data. This input data is a combination of the unstructured type such as text and images and structure obtained from HTML.

Web information extraction and reinforcement learning on the internet are thus among the set of problems that require accurate representations of elements on a web page. Since web pages do indeed provide both structured HTML data and unstructured data in parallel, we explore methods that make full use of such types of data to provide a representation thereof. Web pages are made up of HTML source code and rendered by the browser after applying Javascript code and CSS styles, to finally produce a Document Object Model (DOM) tree structure object. The DOM tree can be interacted with to retrieve all properties and attributes of elements in the web page. We focus our research on representation generating techniques, specifically on models making use of the DOM tree as they are a more information-dense resource then page screenshots or simple source code.

A variety of methods for web element representation have been proposed over the last twenty years, previously using simple bag-of-words and statistical analysis \cite{kosala2000web}, and later on with supervised machine learning techniques such as Support Vector Machines (SVM), kernel estimation, random forest or Naive Bayes \cite{kotsiantis2007supervised}. More recently, deep learning methods have been used including fully connected networks using the attributes of nodes, recurrent networks for text and convolutional networks for images. However, it seems that no previous work has made use of recursive networks on the DOM tree. This work aims to do exactly that.

\section{Related Work}

\subsection{Web Information Extraction}

In this section we present related work in the field of web element classification, for the uses of both web information extraction and for reinforcement learning state- and action representations. The related work on web information extraction shows the scarcity of machine learning techniques applied in this field, whereas the work on reinforcement learning is from very recently and does not make use of the DOM tree structure in the same way as we propose.

Related work in the web page content extraction field is approached in many different ways. \cite{lopez2012using} make use of the DOM tree, but only to calculate a pure content-to-tag ratio. They show this to be a useful method for finding the general "main-content" sections of a webpage where the largest body of text is indeed the content of interest. Although this technique may be applicable for some tasks, it is not precise enough to classify individual elements, or to be generalized to pages where the main content is not a large text body.

Many information extraction techniques use some form of partial content deletion. \cite{gupta2003dom} interact with the DOM tree and run a node-by-node filter over the entire web page. Task-specific layers of the filter may delete a node if it matches the given heuristics, such as an advertisement filter layer deleting nodes based on the URLs the node links to. This technique could be used as a preprocessor for further tasks, but falls short of identifying the leftover nodes as any particular element.

Web pages are not the smallest distinguishable piece of information, and often a page may contain various semantically unrelated elements that do not contribute to the main topic of the page. \cite{cai2003vips} pose that determining the content structure of the web page improves the performance of web information retrieval. Their research shows that users expect certain elements of web pages to be positioned in particular locations, such as navigational menus in the top, advertisements on the sides etc. Because of this user assumption, they propose VIPS, a vision based page segmentation algorithm. This algorithm creates visuals for every node in the DOM tree, and then uses visual segmentation techniques to determine coherent zones, which are finally deduced back to the original element of the web page. Since this VIPS is mostly visual based, it is related but orthogonal to our work.

\cite{spengler2009learning} consider the problem of extracting content from online news web pages. To this end, they first generate a custom dataset of 600 handpicked news pages, and annotate each of the nodes in the DOM tree that contains text or an image. They then propose a classification framework of two different variations: a linear Support Vector Machine and a sequential Conditional Random Field. As input data they take the annotated nodes of the web pages and generate features for each. The features chosen in this paper are highly comprehensive but limited mostly to non-structure information. All available node level features are considered such as an embedding of the text content, attribute values of the nodes, background colours, font size, amount of margin and more. However, the only information regarding the structure of the tree or the context of the node that is embedded in the features, is the tag type of the parent node. Both the SVM and CRF that are proposed by this paper perform well on labels for which very large amounts of data are available and that are relatively simple, i.e. text-only nodes. For more complicated nodes such a "comment section", and nodes that have fewer labelled examples, the performance drops drastically. It is not discussed in their work if the models generalize to other types of web pages then news pages.

\cite{liu2018reinforcement} propose DOMnet, a neural architecture designed to capture spatial and hierarchical structure of a DOM tree. Inside this framework they use a novel DOM embedder, that captures various interactions between web elements. At first all properties and attributes of the element are embedded and concatenated to make an element base embedding. Then, "neighbour" embeddings are created of two types: the first type is a sum of all base embeddings of neighbours that are located within a predefined number of pixels from the element of interest. The second type of neighbour embedding is a concatenation of max-pooling of "local" neighbours. The locality is determined as all tree elements sharing a lowest common ancestor of predefined depth in the tree. Drawbacks of this approach are that both the neighbourhood parameters and spatial radius parameters are hand-picked and thus must be hand-tuned. Furthermore, the second type of neighbours define the context of the element to a specific chosen pattern and do not allow the model to learn where the relevant context of an element is.

Graph Neural Networks are models that capture the dependence of graphs via message passing, proposed originally by \cite{scarselli2009graph} These models may be used for any problem where the data can be represented on a graph, such as classification tasks in molecular biology, modeling diseases, physics systems and data mining. GNNs principally compute state for a neighbourhood of a certain size, but we consider them to be generalised for our use case. Since our problem space is fully restricted to trees, we can make use of the structure of trees which is more informative than that of generic graphs.

As this work was being completed further work was published on GNNs that compute the theoretical limits of their application by \cite{xu2018powerful} showing promising results even for tree structures, and \cite{jia2019dom} propose DOM-Q-Net that uses GNNs as a new DOM embedder for a DOMnet-like reinforcement learning framework.

\subsection{Tree Long-Short Term Memory Networks}

From our overview of DOM tree related information extraction work, we observe that researchers are having difficulties tackling DOM-trees with deep learning techniques and only very recently more interest is arising on applying graph neural networks to trees. We intend to approach the DOM trees by borrowing a neural network technique recently proposed in Natural Language Processing, tree based long short-term memory (Tree LSTMs). We creatively apply this technique to a new problem space, DOM trees instead of NLP, and we will propose novel improvements to Tree LSTMs. 

Natural language models come in three varieties: bag-of-words, sequential models and tree structured models. Bag-of-words models are unordered and thus insufficient for fully capturing the semantics of natural language. Focus in the NLP space therefore lies on the latter two types of models. 

Tree structured language models are built around the syntactic parse trees of sentences, of the dependency tree or constituency tree variety. The appendix in section \ref{apx:parsetrees} explains in more detail where parse trees come from, how they are constructed and what variations exist.

\cite{tai2015improved} search to verify tree structured natural language models perform better than their sequential counterparts. They propose a novel neural network architecture that is a combination of a recursive and recurrent neural network called TreeLSTM.

An ordinary LSTM unit takes an input at a timestep, combined with the output from one previous timestep, whilst keeping an internal memory state. The Tree LSTM architecture is a generalization of such classical LSTMs that allows non-linear sequences to be processed. They implement a new LSTM unit that can process a node at a given time step using outputs and hidden state from multiple previous timesteps instead of just one. Using this new unit, they propose two architecture variations: Child-Sum tree LSTM and N-ary tree LSTM. The first variant allows for operation on a variable number of children in a tree. To compute on a node, it ingests all outputs and hidden states from that node's children. The second variant, N-ary tree LSTM requires the input tree to have a fixed branching factor. Child-Sum tree LSTMs are useful for their flexibility to ingest data, whereas n-ary tree LSTMs fall short in that aspect but allow for fine-grained learning of picking which child's output is relevant.

The Child Sum tree LSTM is well suited to syntactic dependency parse trees, where each node has an arbitrary number of children and contains an input word, and the children of a node are in no particular order. In constituency parse trees, the N-ary model is more applicable, since they are usually binary trees. In these trees, the position of the child (i.e. left or right) is of importance and thus can be modelled correctly.

One of the tasks this new architecture is evaluated on is the Semantic Relatedness. In this task, two sentences are provided, and the model must predict a value that indicates how related the two sentences are based on their dependency parse trees. The authors first embed the words into pretrained GloVe word vectors \cite{pennington2014glove} and thereafter apply the new Child Sum tree LSTM. They show this model outperforms both non-neural-network baselines as well as conventional (bidirectional) LSTM models.

LSTM architecture naturally have a direction due to the nature of how the input sequence is processed by the LSTM unit. \cite{graves2013hybrid} showed that adding a second direction to make a bidirectional network improved performance on problems such as speech recognition by a significant margin, and many other natural language models make use of such bidirectionality. Similarly, \cite{teng2016bidirectional} propose a second direction to tree LSTM. Once again parse trees are the data of interest in this paper, but contrary to \cite{tai2015improved}, here only constituency parse trees are considered. This means that the intermediary nodes in these trees do not represent a word but only a grammatical connection. The only nodes in these trees that represent words are the leaf nodes, and therefore there exists only a feature vector for leaf nodes in this tree, here in the form of a word-embedding. A bottom-up tree LSTM can be applied here since the computation starts at the leaf nodes. The opposite direction computation, top-down, however is not initially possible and not as intuitive as in linear sequences. Since constituency trees have no features for non-leaf nodes, any computation starting at the root will not encounter features until the end of computation. It is therefore expected that such a computation will not make much sense.

To tackle this issue and ultimately create a bi-directional TreeLSTM, the authors propose a "head-lexicalization" method that will provide feature vectors for every node in the tree. This method produces a "head-word" for a node, that is estimated to be a representative word for the sub-phrase capture by this node's children.

Once these representative "head-words" are generated for all nodes, a top-down method can be implemented on the tree. The top-down method in this paper is a LSTM over the words in the path from the root to the node of interest. Since the words are completely sequential in this direction, it is thus highly similar to a linear LSTM. 

The paper shows their results on a two tasks: sentiment classification, and question-type classification. Their results show that this new model outperforms the known state-of-the-art at the sentiment classification task, and performs similar to the state-of-the-art on the question type classification task. In both cases the model size, defined as number of parameters, is only up to 3 times larger than single-direction LSTMs.

\cite{miwa2016end} show the use of tree LSTM for extracting semantic relations between entities of text phrases. At first, named entities are predicted for all word embeddings in a sentence using a conventional linear bidirectional LSTM in combination with a multilayer perceptron (MLP) and a decision layer. To determine the semantic relation between a pair of such named entities a second model is used, that can exploit the syntactic parse structure of a sentence. To this end, the authors propose a variation of tree LSTM where computation is only done on the minimal subtree that includes the shortest path between the nodes representing both entities of the pair. This subtree is essentially just the shortest path between the pair, but rooted at the lowest common ancestor of the two nodes.

A bottom-up representation is calculated on this special subtree by using a Child-Sum-like tree LSTM. The second, or top-down, direction is a sequential LSTM that computes a representation for each of the two leaf nodes. This input sequence for this top-down components is the path from the root of the subtree to the leaf node, for both leaf nodes. The bottom-up representation, and both top-down representations are concatenated together to represent the relation between the two entities. Finally, a MLP and a decision layer predict a relationship label from this relation representation.

The authors conclude from their variety of experiments, such as the ACE05 relation prediction task, that their model outperforms previous non-tree-LSTM models on relation prediction. One drawback of this method is that the performance improvement is valid specifically for the relationship task, which limits its application to only pairs of inputs.

From our overview of the various implementations and additions to tree LSTM we recognise their success in achieving state-of-the-art performance in a variety of tasks by capturing the additional information contained in the structure of the input data. Even though these models were only evaluated on language parse trees, we estimate that since the underlying structure is the same, we may use implementations similar to these models on a new domain of DOM trees and produce efficient representations of subtrees. 

In this report we propose the following contributions:
\begin{itemize}
\item Application of tree LSTM models to a new domain; the web.
\item Novel bi-directional extension to tree LSTM which computes context-aware representations of subtrees
\item Performance analysis of our extension on a web classification problem.
\end{itemize}

\section{System Model \& Problem Statement}

Tasks such as web information extraction and reinforcement learning on the web need comprehensive representations of web page elements to perform efficiently. Current methods for building such representations fail to fully utilise the context of the elements and therefore are not efficient. We aim to adapt a technique from NLP to generate comprehensive, efficient representations for elements in web pages. Since web pages can be represented as tree structures, our ultimate goal is to extend this method for generating representations of a subtree in any domain.

\subsection{DOM Tree}

The Document Object Model (DOM) \cite{whatwgDOM} is an API that is used to represent an HTML web page as a tree structure, where each node is an object representing an element of the document. The main difference between the HTML code source tree and the DOM tree is that the DOM tree objects also contain information about the rendered state of the page, i.e. after any CSS styles or Javascript code have been applied and the browser has rendered the web page to the state that the user observes it in. Furthermore, the DOM tree contains by design the structural relationships between elements, encoded in the structure of the tree.
A node of the DOM tree is thus an element on the web page, where the contents of that element are in the subtree rooted at the element node, and elements that are in the remainder of the tree are the context of that element. Elements that are close neighbours in the DOM tree may be in spatial or logical proximity in the web page. Until now there are no methods of processing a DOM tree into a neural network, and thus we want to generate lower-dimensionality representations of such trees. We will make use of the DOM tree to reason about the properties and attributes of each element for use as features, and we compute our representations over the tree structure.
In our model, the input data is such subtrees of the DOM tree, and the goal is to compute a reduced-dimension context-aware representation for that subtree. To show a potential usage application of such representations, and to evaluate their precision, we will use these representations to predict the semantic meaning of the element.

        \begin{figure}[h]
        \centering
                \begin{subfigure}{0.5\textwidth}
                \centering
                    \includegraphics[width=0.9\linewidth]{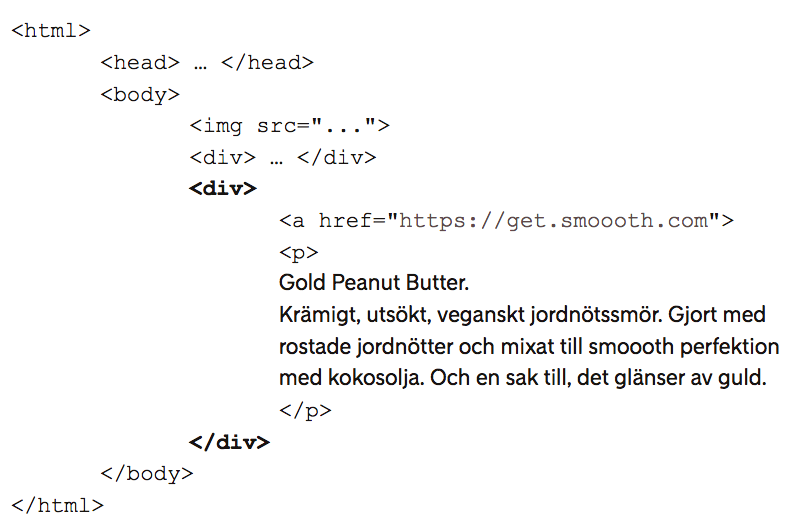}
                \end{subfigure}%
                \begin{subfigure}{0.5\textwidth}
                \centering
                    \includegraphics[width=0.7\linewidth]{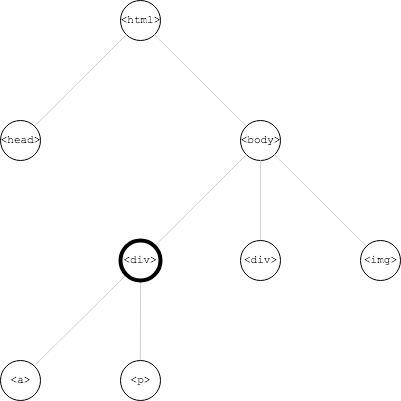}
                \end{subfigure}
                \caption{A simple HTML page source code and a simplified view of the corresponding tree.}
                \label{fig:HTMLcodeAndTree}
        \end{figure}

\subsection{Recurrent Neural Networks}
We will propose an application of tree LSTMs to such DOM trees and a novel extension to this architecture that will capture context in a more effective fashion. To support our new architecture proposal, we review some aspects of the LSTM neural network type.

\subsubsection{The LSTM unit}
\label{sec:LSTMUnit}

The Long Short Term Memory unit is a particular variant of a recurrent neural network (RNN) architecture, originally proposed by\\ \cite{hochreiter1997long}. LSTM models, like RNNs models, are a form of Artificial Neural Networks (ANN) that are Turing-complete because of their feedback connections \cite{siegelmann1995computational}. Unlike ANNs that only allow for the processing of one data point at a given time, recurrent networks can process entire sequences of data, such as speech or video. One drawback of RNNs is that they are prone to the vanishing gradient problem, which makes detecting long term dependencies in time series difficult. LSTMs were specifically designed as an improvement to classical RNNs to address this vanishing gradient problem. Contrary to traditional RNNs, LSTMs employ the use of an internal memory cell to contain long-term dependencies, and a set of gates to modulate the information passing from the memory cell, the previous state, and the current input. Equation set \ref{eq:vanillaLSTM} describes the inner workings of a common LSTM unit.

\begin{equation}
\begin{split}
i_{t} 		&= \sigma(W^{(i)}x_t + U^{(i)}h_{t-1} + b^{(i)}) \\
a_{t} 		&= tanh(W^{(a)}x_t + U^{(a)}h_{t-1} + b^{(a)}) \\
o_{t} 		&= \sigma(W^{(o)}x_t + U^{(o)}h_{t-1} + b^{(o)}) \\
f_{t} 	    &= \sigma(W^{(f)}x_t + U^{(f)}{h}_{t-1} + b^{(f)}) \\
                                            \\
c_{t} 		&= i_t \odot a_t + f_t \odot c_{t-1} \\
h_{t} 		&= o_t \odot tanh(c_t)
\end{split}
\label{eq:vanillaLSTM}
\end{equation}

In these equations, $a_t$ is the activation of the input at timestep $t$ and the hidden state of the previous timestep $t-1$. This activation is modulated by input gate $i$, and then the internal memory cell is modulate with the forget gate $f$. Then, the internal memory cell is updated by the sum of the input activation and the memory cell contribution, and finally the output of the unit is the hidden state, which is an activation of the memory cell that is modulated by the output gate. It is clear from these equations that only the hidden state of one previous time step is considered when computing the current hidden state.

\section{Our Solution}

To generate our context-aware representations for subtrees that will serve as a lower dimensional efficient representation of a tree, we propose a dual contribution. First we apply a variant of tree LSTM borrowed from NLP to DOM trees and evaluate its performance for a classification task, and secondly, we propose a novel extension to tree LSTM.
In our model of the system, we assume that each node in the tree has a feature vector that comprises the features local to the node, i.e. these features do not contain information regarding the structure in which the node resides. We intend to generate representations for given subtrees in a bidirectional manner. 

\subsubsection{Bottom-Up Tree LSTM unit}

The first component of our solution is an application of Child-Sum tree LSTM as proposed by \cite{tai2015improved}. We will use this method to compute a representation of the subtree using its local context, where in the frame of reference of DOM trees, the local context would be the contents of the element corresponding to the subtree.
Equation set \ref{eq:childsumLSTM} defines the child-sum tree LSTM unit:

\begin{equation}
\begin{split}
\widetilde{h_j}  &= \sum_{k \in C(j)} h_k \\
i_{j} 		&= \sigma(W^{(i)}x_j + U^{(i)}\tilde{h}_j + b^{(i)}) \\
a_{j} 		&= tanh(W^{(a)}x_j + U^{(a)}\tilde{h}_j + b^{(a)})   \\
o_{j} 		&= \sigma(W^{(o)}x_j + U^{(o)}\tilde{h}_j + b^{(o)}) \\
f_{jk} 	    &= \sigma(W^{(f)}x_j + U^{(f)}{h}_k + b^{(f)})       \\
                                                                 \\
c_{j} 		&= i_j \odot a_j + \sum_{k \in C(j)} f_{jk} \odot c_k \\
h_{j} 		&= o_j \odot tanh(c_j)
\end{split}
\label{eq:childsumLSTM}
\end{equation}

The equations of the tree LSTM unit are very similar to the classical LSTM units as presented in section \ref{sec:LSTMUnit}, except for the aspects that handle the hidden state and internal memory cell from the previous timesteps. In the bottom-up methodology, the previous timesteps of the data correspond to the children of the node $j$ of the current timestep, defined by the class $C(j)$. In the previous timestep, each of these children nodes have produced a hidden state $h_j$, and a memory cell $c_j$. In this variant, we sum all child-hidden states into $\widetilde{h_j}$ (equation \ref{eq:childsumLSTM}.1), and use this summed hidden state for the input, activation, and output gates. We need to modulate the contribution of each child's memory cell individually, and therefore a forget gate $f_{jk}$ is made for each node in $C(j)$.

\begin{figure}[h]
    \centering
    \includegraphics[width=0.8\linewidth]{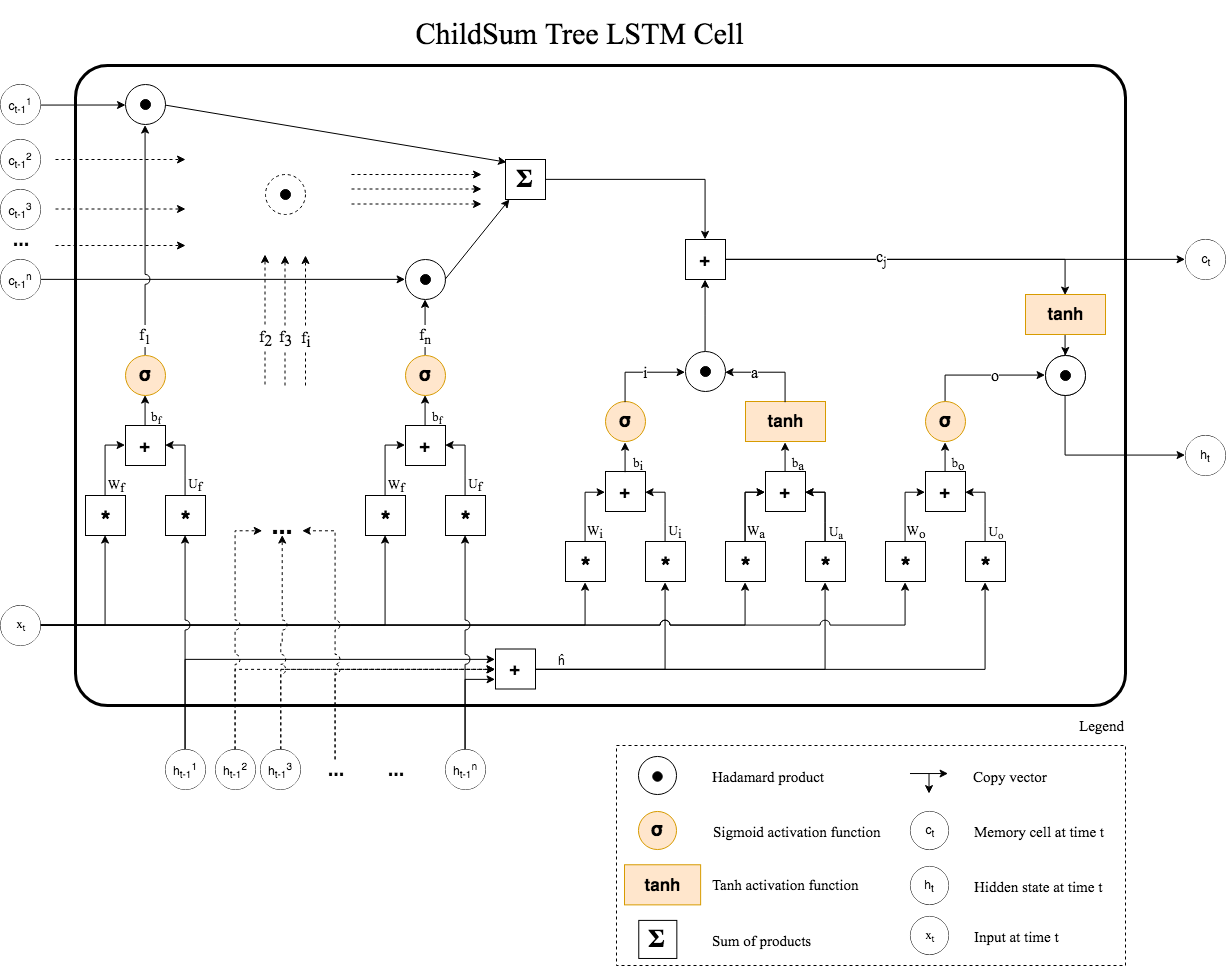}
    \caption{The innerworkings of a Child-Sum Tree LSTM Unit}
    \label{fig:treeLSTMunit}
\end{figure}

\begin{figure}[h]
    \centering
    \includegraphics[width=.7\linewidth]{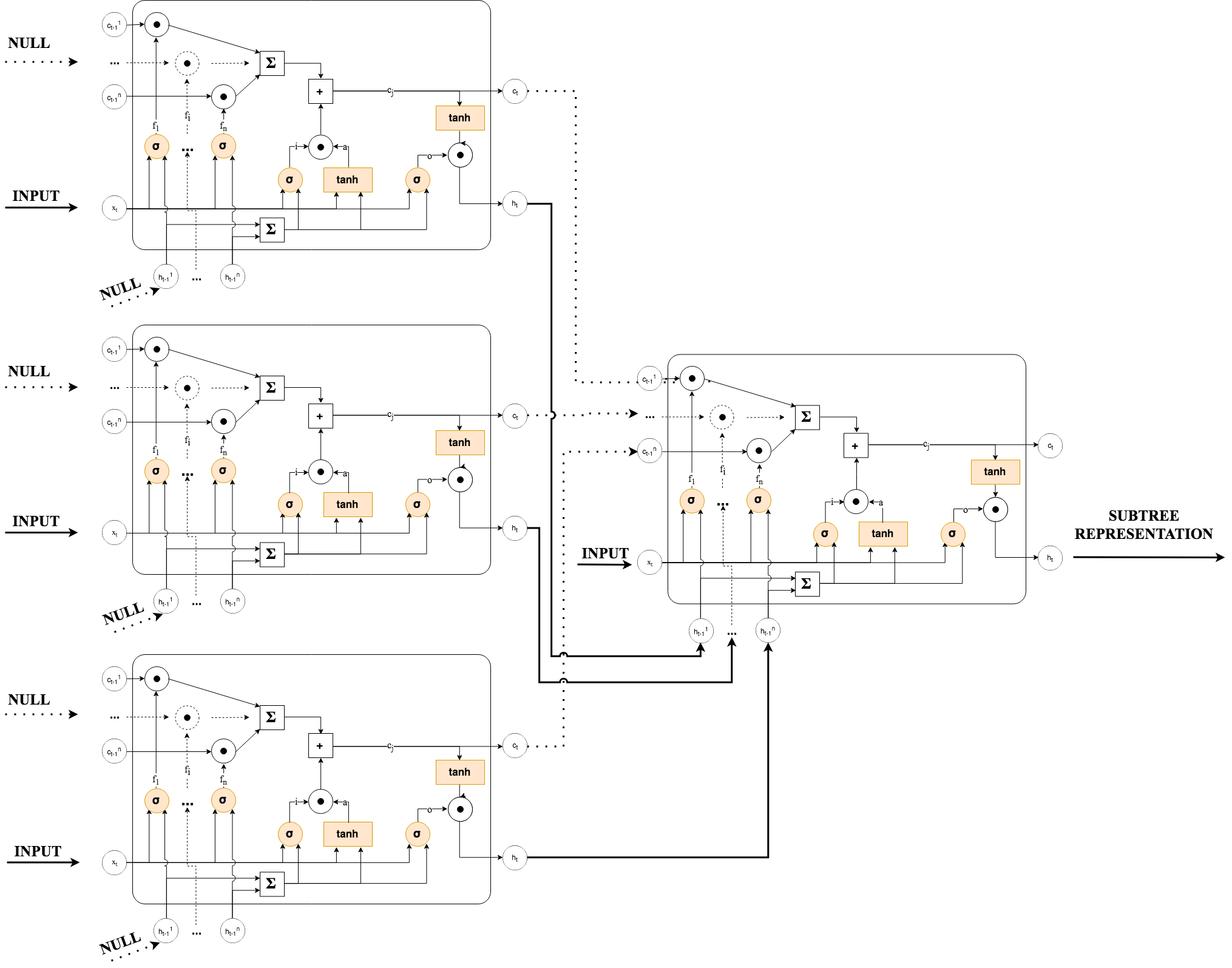}
    \caption{An "unrolled" overview of the interconnections between Child-Sum Tree LSTM timesteps.}
    \label{fig:treeLSTMoverview}
\end{figure}

Finally, the hidden state and memory cell of the current timestep are completed using the last two equations. The leaf nodes of a tree structure have no children, so their computation is solely based on the input feature vector activation. A visual reference to the inner workings of the tree LSTM unit is provided in figure \ref{fig:treeLSTMunit}. Figure \ref{fig:treeLSTMoverview} shows the un-rolled view of the application of such a LSTM unit to a small tree.

\subsection{Bottom-Up Tree LSTM method}
\begin{figure}[h]
    \centering
    \includegraphics[scale=0.6]{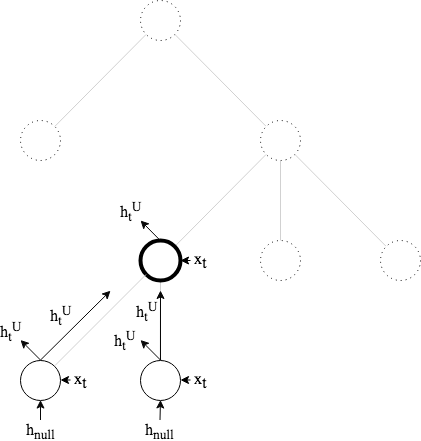}
    \caption{Visual representation of the Bottom-Up tree LSTM recursion}
    \label{fig:methodBU}
\end{figure}

Using the bottom-up tree LSTM cell we compute the final hidden state, produced on the root of the subtree, which is the representation of that subtree. It is depicted in figure \ref{fig:methodBU} by the top most $h_{t}^{U}$.

In order to obtain information from the non-local, or global context of the subtree, we have to consider the position of the subtree in the full tree. For example, the bottom-up component may be very accurate in computing a representation of a web element such that it can be classified as a "add-to-cart" button, but it may not distinguish between what particular add-to-cart button. On many product web pages, we can find one such primary add to cart button, and a set of secondary add-to-cart buttons that are for recommended products elsewhere on the page. The context information that will determine which element, and thus which subtree, is which, is the position of the element in the tree.

\subsection{Top-Down Tree LSTM on features}
\label{sec:methodTDfeatures}

\begin{figure}[h]
    \centering
    \includegraphics[scale=0.6]{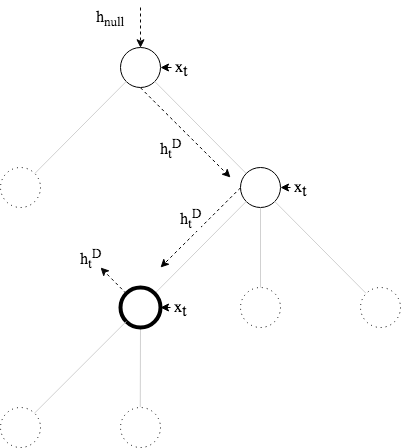}
    \caption{Visual representation for the Top-Down tree LSTM recursion over node-local features.}
    \label{fig:methodTDfeatures}
\end{figure}

To this end, we propose a second "direction" to our architecture, that encodes a representation for the path from the root of the full tree to the root of the subtree. This tree LSTM unit will compute on the sequence of nodes on this root-subtree path, and will receive the feature vector of each node as the input. Since the computation goes down the tree, the LSTM unit receives the hidden state and memory cell of the parent of the node in each timestep. Since this component of the model computes only on a path, the top-down tree LSTM essentially functions like a regular linear LSTM. Again, the hidden state and memory cell of the first element in the computation, in this case the root of the full tree, are computed purely on the root's features, since it has no parent. The hidden state produced by the last timestep of this component, which is the root of the subtree, we call $h_{t}^{D}$, as shown in figure \ref{fig:methodTDfeatures}.

The concatenation of our bottom-up and top-down tree LSTM component outputs, $h_{t}^{U}$ and $h_{t}^{D}$ respectively, is the representation of the subtree in this architecture. This model  is our first variant of Tree LSTM, "Bidirectional tree LSTM on features".

\subsection{Top-Down tree LSTM on context embeddings}
\label{sec:methodTDembeddings}

To fully capture all context for a subtree, we introduce a novel tree LSTM architecture extension. For some tasks, there is a significant amount of information in the context of a subtree. We estimate that tasks such as web element classification may benefit from the semantic meaning of such context. Our novel tree LSTM architecture is built in a similar, but distinct method to our previously discussed "Bidirectional tree LSTM on features". The bottom-up component of this method remains the same as proposed before, but the following novel top-down method replaces the previous version. 

\begin{figure}[h]
    \centering
    \includegraphics[scale=0.6]{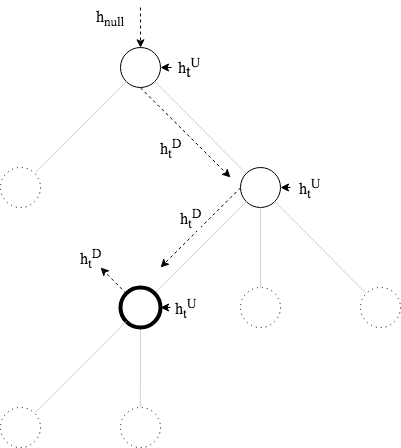}
    \caption{Visual representation for the Top-Down tree LSTM recursion over bottom-up computed embeddings.}
    \label{fig:methodTDembeddings}
\end{figure}

First, we compute the bottom-up representations for all nodes in the full tree that the subtree is contained in. To this end, we employ the same bottom-up tree LSTM unit and method as discussed before. In the context of this top-down method, we call these preliminary representations of the nodes "node embeddings". Second, we compute a tree LSTM over the path from the root of the full tree to the root of the subtree, similarly to section \ref{sec:methodTDfeatures}, but use the "node embedding" for each node as an input to the tree LSTM, instead of using the node's local features. The hidden state and memory cell are propagated as before. This top-down component now propagates the embedding of the whole tree down to subtree, in effect passing the context to the node. Figure \ref{fig:methodTDembeddings} shows a visual representation of this component, where the notable difference to figure \ref{fig:methodTDfeatures} is in the input for each node, here $h_{t}^{U}$ instead of $x_t$ like in the previous version.

Finally, this model outputs the concatenation of $h_{t}^{U}$ and the fully-context-aware $h_{t}^{D}$ as the representation for the input subtree. This second variant on tree LSTM we name "Bidirectional tree LSTM on embeddings".

All three methods above will generate representations for subtrees, where our novel method will produce the most context-aware representation, that may be used in a variety of information extraction and reinforcement learning tasks.

\section{Evaluation}

This section will discuss the experiments that were conducted to evaluate the models. 
It is difficult to appreciate the correctness of a generic representation of a subtree, and as such we add two tasks that aim to use the representations produced by our tree LSTM models to then classify the subtree. The performance evaluation can then be done on the error of the predicted classification against the ground truth.

To benchmark the base implementation of our bottom-up tree LSTM, we compare our model's performance to those proposed by the literature which uses a task with known benchmarks. This task is Task 1 from SemEval 2014 which focuses on semantic relatedness of phrases, based on the grammatical structure of the phrases.

After establishing our model's performance against a known dataset, we evaluate its effectiveness, and the effectiveness of our novel extension, on our own task and dataset. Our task is to predict the semantic meaning of elements in a webpage using the underlying DOM tree. We produce a new dataset comprised of 25'000 web pages that were labelled by hand.

\subsection{Semantic Evaluation}

The Semantic Evaluation 2014 Task proposes the data set called Sentences Involving Compositional Knowledge (SICK) \cite{marelli2014semeval}. The goal of the task is to predict the degree of semantic relatedness of two given sentences.
The data set includes 10'000 sentence pairs specifically produced for this task. The sentence pairs are rich in lexical, syntactic and semantic phenomena. Each pair of sentences was manually annotated as an average of the evaluations of ten human experts in terms of their semantic similarity on a scale of one to five (not related - highly related). For each sentence in the dataset, the grammatical structure of the sentence is also provided. We only evaluate our Monodirectional Tree LSTM on this dataset, since the task only requires predictions on entire trees, not subtrees. Our bidirectional models contribute mainly in tasks where the subtree is the focus. Our implementation of Monodirectional bottom-up tree LSTM on this dataset and task performs very similar to the results presented by \cite{tai2015improved}, where they produced a Mean Squared Error of 0.2532 using \textit{Dependency Tree LSTM} vs. our implementation's 0.2621. This difference is rather small and may be accounted for by further optimisation of hyper parameters that the authors may have performed that we are unaware of. These performance figures indicate that our implementation is correct.

\subsection{Klarna Webpages}

\subsubsection{Base Data}

Since Klarna is interested in the task of web page element classification, we were able to allocate some means in order to generate a dataset. The goal of the dataset is to be a collection of web pages, with labels on a small subset of the elements in the web pages.. The dataset also has to be representative of some factors which we may encounter in the real world, such as country of origin, written language, as well as the owner of the website. Our particular problem is about classifying web elements of product pages on web shops. A web shop is owned by a merchant, and is assigned to a country.

To tackle this problem, Klarna has created a dataset consisting of several thousand labelled webpages. First, a list of URLs to scrape was generated. This list was handcrafted to the following specifications:

\begin{enumerate}
\item 5 product pages per merchant
\item Between 400 and 2800 merchants per region
\item Regions: US, UK, SE, DE, NL
\end{enumerate}

The complete dataset consists of 25,000 product pages. Each of these product pages is then manually annotated using an in-house Klarna tool. Each annotation consists of 5 labels on the page. Each label is assigned to the minimal element that fully answers the requirements for that label. The following labels were used, with their definitions below:

\begin{enumerate}
\item Main product name
\item Main product price
\item Main product add-to-cart button
\item Main product main product picture
\item Merchant go-to-cart button (cart)
\item Subject node
\end{enumerate}

The \textsl{Main product} on the page, as its name suggests, is the primary product that the product page is trying to sell, contrarily to the other products on the page which may be derivatives, suggested other products, or completely unrelated upsells. The name of the product is the human-expert chosen text phrase that seems to describe the product best on the page. The main product picture is the largest picture of the product, that is visible when the page is loaded. The price and the add-to-cart button are self explanatory. The merchant go-to-cart button is a button or link that, when clicked, takes the consumer to a new section or page which shows all the items that they have put in to the checkout-basket for the moment.
For each of these, the "minimal element" is defined by the HTML tag that is the lowest in the tree that encompasses fully the requirements for the element. When the human expert annotates such a minimal element with its label, the tool sets a custom attribute on the HTML element which contains the chosen label. After having chosen all 5 elements for a page, the tool saves the web pages including these labels to a database.

\subsubsection{Data Processing}

At this stage, we have a database filled with web pages obtained from a reasonable distribution. These web pages are stored in the MHTML format, which allows us to not only inspect their source code, but also obtain the full DOM tree, which represents how the web page looked once all CSS styles and JS logic were applied. This DOM tree allows us to have a more feature-complete look at the web page. For example, for each element in the page we can obtain the coordinate location of where the element was located and the width and size of the element.

From inspecting the dataset, we calculate that the average web page in our dataset is made up of around 2000 HTML tags, or DOM nodes. Since we have labels on 5 nodes of the page, only 0,25\% of the data in a web page is labelled. Each node can be considered to be a subtree of the full DOM tree of the web page. If we want to make a context aware representation for any subtree, than we need to consider, on average, all 2000 nodes of the page. If we do take all subtrees, cardinality of the set of subtrees with no label would be almost 2000 times as large as the cardinality of each of the sets of subtrees with a label. It has been shown that model performance may degrade significantly when training on an unbalanced dataset \cite{mazurowski2008training}. To make our dataset balanced, we uniformly at random subsample by 0,25\% the collection of subtrees with no label, and annotate these elements with the "negative class" label.

For a given number of web pages, we now thus create a dataset that contains $n$ subtrees of interest, where $n = $ number of classes $\cdot$ number of web pages.
Our model relies not only on the structure of the web page, but also on the features of each node in the tree itself. Therefore for each node we create a feature vector of fixed size that contains a subset of all possible information local to that node. These features vectors will serve as the input $x_t$ for the nodes in the tree LSTM models.

We have chosen to leave the optimal feature engineering problem outside of the scope of this research. The following method is used to generate a basic feature vector for each node simply due because they were straightforward to obtain. First, from the rendered DOM tree, we extract these attributes of the node:
\begin{enumerate}
\item X \& Y coordinates of the top-left corner of the element
\item Width of the element in pixels
\item Height of the element in pixels
\item Number of bitmap images (png, jpg, gif) contained in the element
\item Number of vector images (svg) contained in the element
\item Font size
\item Font weight
\item Visibility
\item Active status
\item Tag type
\end{enumerate}

Items [1,2,3,4,5,6,7] are self explanatory and used as an unnormalized scalar in the final feature vector. The visibility of the element is extracted from its rendered state, and may be one of the following 6 options: [hidden, visible, collapse, inherit, initial, unset]. These are the CSS visibility options for an HTML element. This attribute is mapped to a single scalar that has as value, the index of the option it is.

The active status is a boolean that indicates whether the element is interactable or not. In a webpage, an interactable element is any element that is a button, link, or has a functionality that triggers when clicked on. This boolean is mapped to a 0 or 1 scalar. Lastly, the tag type is a one-hot encoding of the tag of the element. The element's tag may be \texttt{body}, \texttt{div}, \texttt{p}, \texttt{a}, \texttt{span} etc. If the tag is not in the set of 59 most commonly seen tags, then we set it to \texttt{UNK}. This thus creates a 60 size one-hot sparse encoding of the tag.
All these vectors and scalars concatenated make a one dimensional, 70 size vector as the input feature vector of the node.
Further use of the models proposed in this research may spend new effort on feature engineering, and including embeddings of information high-density features such as a representation of the text contained in the element, the HTML ID and Classes of the element, and even the image in the element. All of those text features can be approached by any NLP model to generate an embedding, and the images may be reduced to an embedding through a state-of-the art CNN for example.

\subsubsection{Data augmentation}
Another element of interest was determined to be what we call the "subject node". In the context of product pages, we established the subject node to be the minimum element that includes the product name, price and picture. More specifically, it is the lowest common ancestor of the subtrees of the name, price and main picture. In order to annotate the subject nodes of the web pages, we implemented a method that finds the lowest common ancestor for a given collection of nodes, This method was then executed on each of the web pages in our dataset, and produced one more label on that web page, increasing the number of labelled nodes from 5 to 6.

\section{Models}

Four distinct neural networks, and fuctionality to train them, were implemented in order to compare their performances. Each of the networks has the same task of taking a subtree as input to the model, and producing a classification prediction of the label of that subtree. The subtrees are provided in the context of their full tree, so that the networks have complete access to all nodes of the tree, and their attributes.

\subsection{Classification Layer}

In order to fairly compare the models, and because their tasks are the same, the output layer and the cost function of all networks are the same. The output layer consists of 7 units, where each unit represents the negative class or any of the 6 positive classes. The  prediction of the model is a softmax function applied over the output layer. The softmax function (Equation \ref{eq:softmax}) is a generalized multiclass alternative to the logistic activation function, that is well suited for classification problems. We use the stable softmax function for numeric stability.

\begin{equation}
    \texttt{softmax} (\mathbf {z} )={\frac {e^{z_{j}}}{\sum _{k=1}^{K}e^{z_{k}}}} \text{ for } j = 1, ..., K \text{ and } {\displaystyle {\mathbf {z}}=(z_{1},\ldots ,z_{K})\in \mathbb {R} ^{K}}
\label{eq:softmax}
\end{equation}

In our classification task, $K = 7$, as the last layer contains 7 units. The final prediction of the network is then the index of the unit with the highest prediction value, implemented as the argmax function.

The loss function for all networks is the average negative log likelihood (NLL, or $H$ as in Equation \ref{eq:nll}) of the softmax argmax prediction $p_i$ and the ground truth $y_i$ for datapoint $i$ in the training set. The combination of the softmax function for prediction combined with the NLL is often referred to as Categorical Cross Entropy Loss in the deep learning community. Using this loss function we can calculate the error for each datapoint and propogate it throught the model to update the weights.

\begin{equation}
    H(y,p) = - \sum_i y_i log(p_i)
\label{eq:nll}
\end{equation}

\subsection{Model training}

\subsubsection{Dataset Split}
The dataset was split in a stratified fashion, conserving the distribution of countries of the web shops, into a train/validation/test split, respectively 64\%/20\%/16\% of the whole data set. The test set was not used until the very last moment to produce the final performance scores reported in this report as such. The validation set was used during training for hyperparameter tuning.

\subsubsection{Optimizer}

All networks were trained on the training set with Backpropagation through time (BPTT)\cite{mozer1995focused} and Backpropagation through structure (BPTS) \cite{goller1996learning} for the LSTM layers, and the overall optimizer for the entire networks was always Stochastic Gradient Descent (SGD).

In order to minimize the variables considered in the results, training was always done with a mini-batch size of 50, and a learning rate of 0.0025, which were established through hyperparameter tuning early on.

\subsection{Model versions}

\subsubsection{Fully Connected layers only}
\begin{figure}[h]
    \centering
    \includegraphics[scale=0.3]{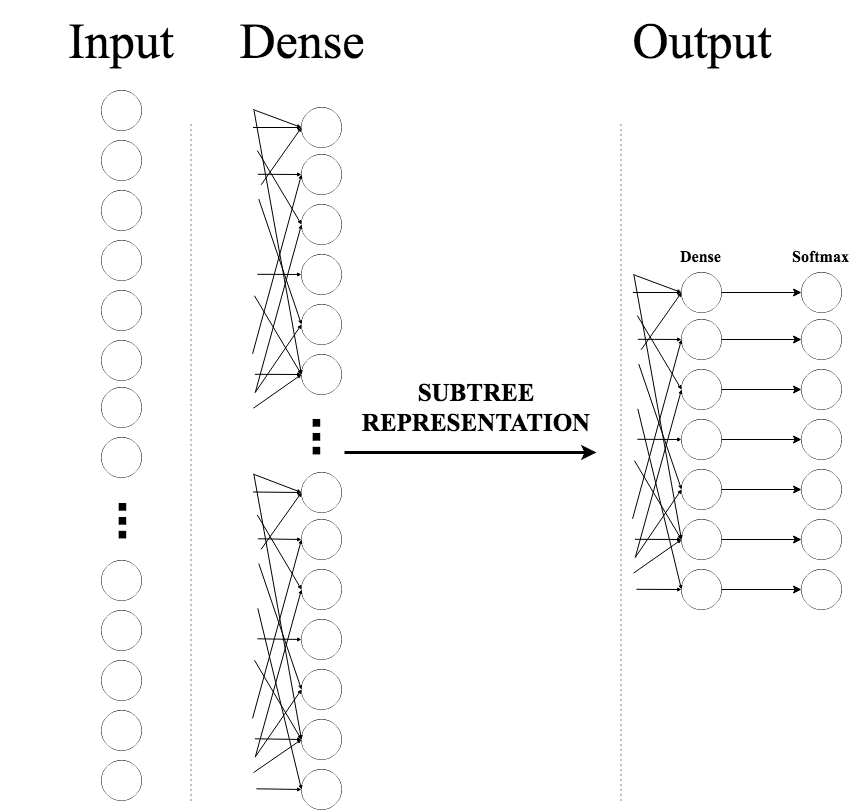}
    \caption{An overview of the Fully Connected layers model.}
    \label{fig:networkFC}
\end{figure}

Our first network is a very basic ANN model that serves as a baseline in terms of both performance and training time to our other experiments. In this network, we only consider the root node of the subtree that is provided as input to the model. The feature vector of this root node is the input to the network. The model then consists of 2 fully connected hidden layers; the first is a 150 units layer, and the second layer is the output layer of 7 units. Figure \ref{fig:networkFC} shows an overview of what the implementation of this model looks like.

\subsubsection{Monodirectional Bottom-Up tree LSTM}
\begin{figure}
    \centering
    \resizebox{\textwidth}{!}{%
    \includegraphics{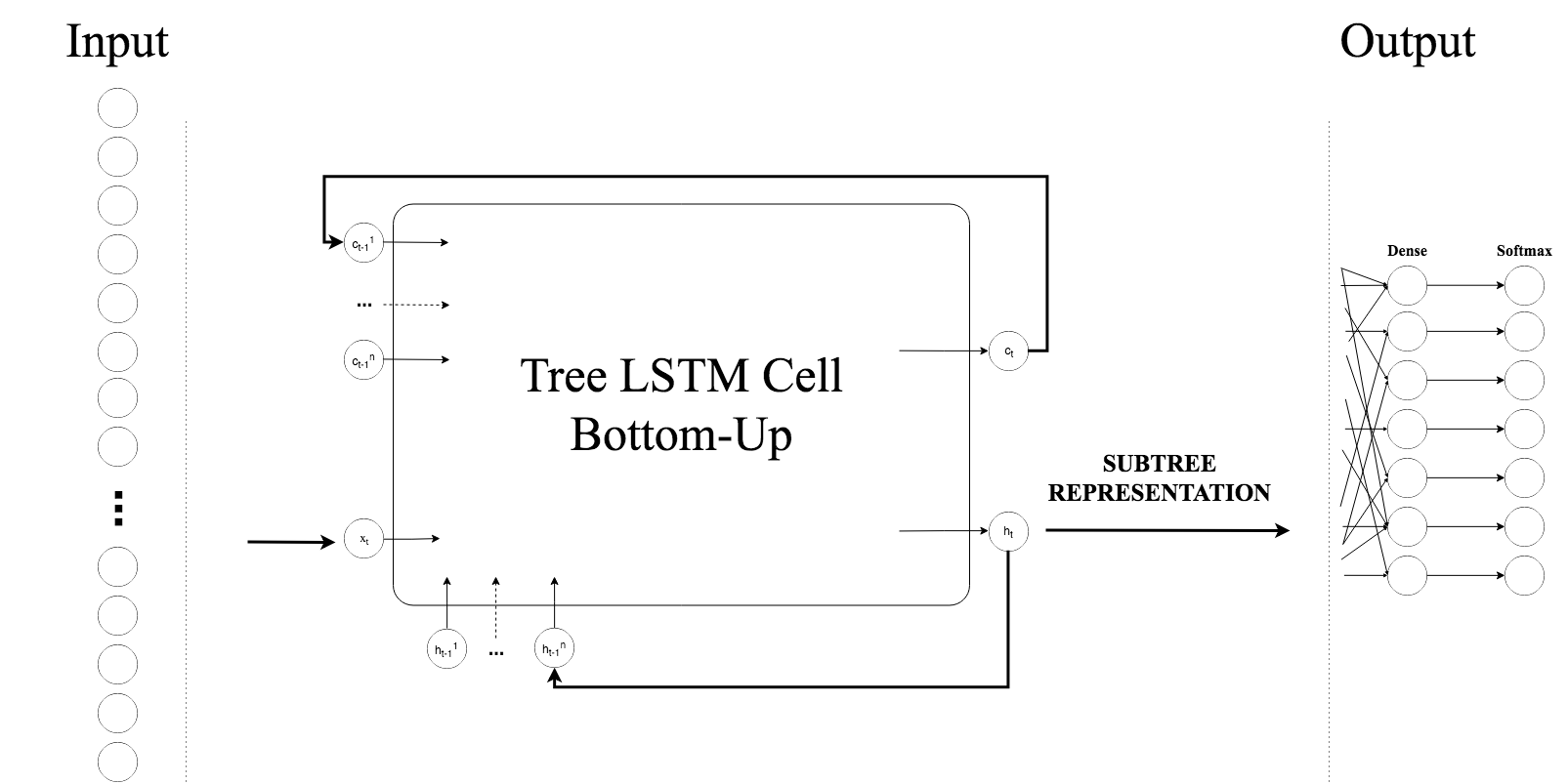}
    }
    \caption{An overview of the monodirectional, bottom-up tree LSTM model.}
    \label{fig:networkMono}
\end{figure}

As described in the model section of this report, this monodirectional, bottom-up architecture stems originally from NLP methods and we are the first to apply it to DOM structures. In this model, we created an LSTM layer that outputs a vector of size 150. Inside the layer is a tree LSTM unit, of the child-sum variety. For a given subtree input, we recursively apply the tree LSTM unit on all of the children. The hidden state embeddings of all children are summed and used in the input, activation and output gates. Furthermore, for each of the children we create a forget gate that modulates it's memory cell contribution. The input for a node at timestep t is the feature vector of the node.
Once the recursion is complete, we have generated a bottom-up-computed hidden state and memory cell for the given subtree. The hidden state is our context-aware representation for the subtree. In order to benchmark its performance, we feed it into the output layer of size 7 like all networks. Figure \ref{fig:networkMono} shows an overview of what this model configuration looks like.

\subsubsection{Bidirectional model using Top - Down root to subtree path}
In order to extract information from the context of the given subtree, we add a secondary LSTM direction to our model. In this architecture the primary direction is a bottom-up tree LSTM, exactly as described in the previous paragraph. The secondary direction is a tree LSTM that is run on the path from the root of the full tree in which the subtree is located, to the root of the subtree itself, as described in Section \ref{sec:methodTDfeatures}.

\begin{figure}
    \centering
    \resizebox{\textwidth}{!}{%
    \includegraphics{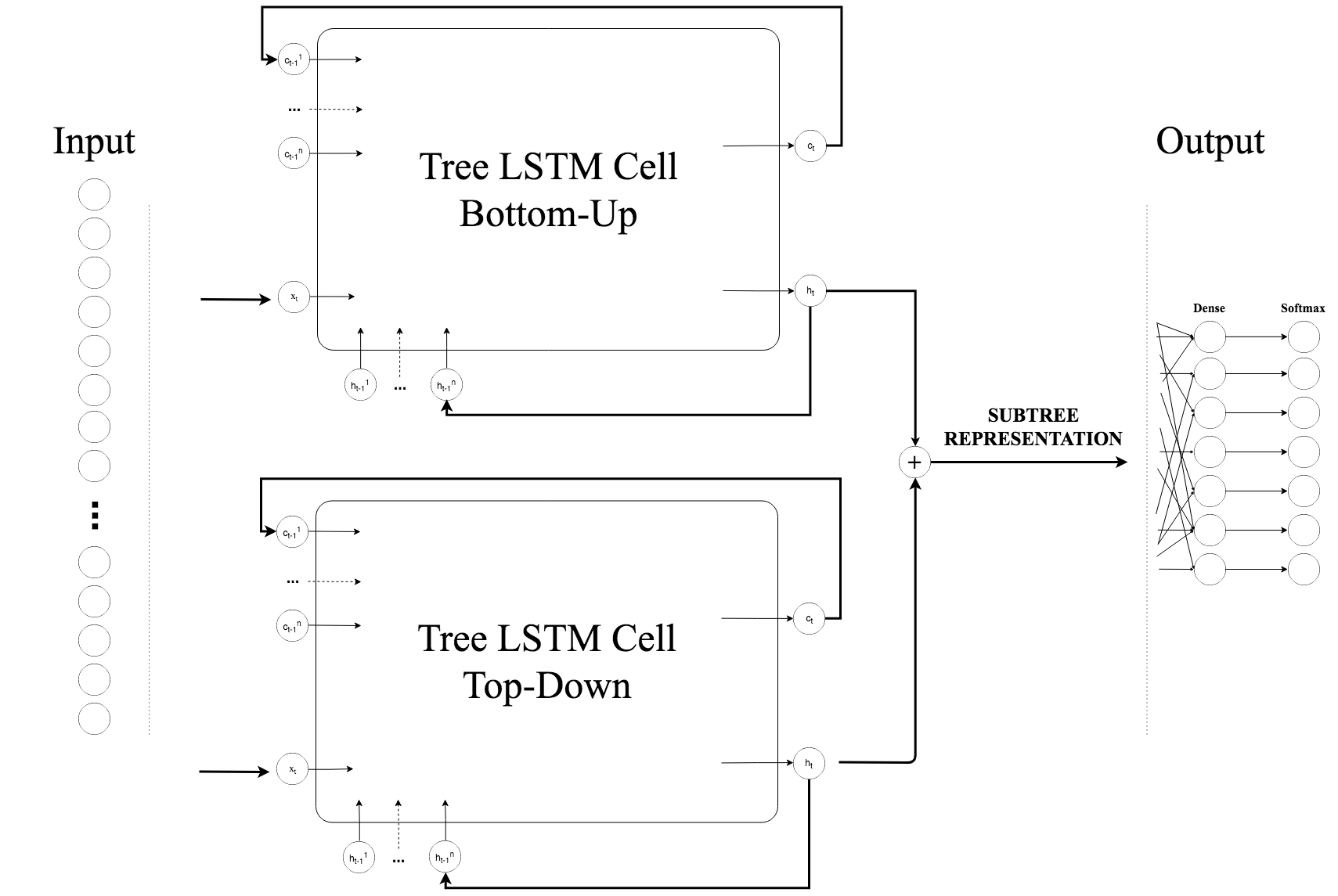}
    }
    \caption{An overview of the bidirectional tree LSTM models.}
    \label{fig:networkBidir}
\end{figure}

This secondary top-down tree LSTM uses its own kernel and is trained in parallel to the bottom-up component. The hidden state of the bottom-up component and the hidden state of the top-down component, are concatenated to provide a context-aware representation of the subtree. The size of this representation is the sum of the sizes of the two component representations, and thus is 300. 
As we did with the other proposed architectures, the representation is fed directly into the output layer to gauge performance. Figure \ref{fig:networkBidir} shows an overview of what this model configuration looks like

\subsubsection{Bidirectional model using Top - Down on full tree embeddings}

As the main goal of our research is to exploit structure in order to produce fully context-aware representations for subtrees, we propose a completely novel method that uses bottom-up embeddings of a tree to pass context to a subtree, as described in Section \ref{sec:methodTDembeddings}. Once again, the two tree LSTM components of the network generate a 150 sized representation for the subtree that is concatenated into a 300 sized representation of the subtree. Finally, like the other architectures, this representation is fed directly into the output layer to gauge performance.

We summarise the memory usage of each model in Table \ref{tab:memoryusage}

\begin{table}[h]
\centering
\resizebox{\textwidth}{!}{%
        \begin{tabular}{|l|r|r|}
        \hline
        \rowcolor[HTML]{EFEFEF} 
        Architecture Type                   & \multicolumn{1}{l|}{\cellcolor[HTML]{EFEFEF}\begin{tabular}[c]{@{}l@{}}Avg. number of params\\ to store per input\end{tabular}} & \multicolumn{1}{l|}{\cellcolor[HTML]{EFEFEF}\begin{tabular}[c]{@{}l@{}}Avg. memory required\\ per mini-batch\end{tabular}} \\ \hline
        Fully Connected                     & 10650                                                                                                                           & 0,00213 GB                                                                                                                 \\ \hline
        Monodirectional BU                  & 7 425 600                                                                                                                       & 1,485 GB                                                                                                                   \\ \hline
        Bidirectional: BU + TD (features)   & 8 751 000                                                                                                                       & 1,750 GB                                                                                                                   \\ \hline
        Bidirectional: BU + TD (embeddings) & 267 006 000                                                                                                                     & 53,401 GB                                                                                                                  \\ \hline
        \end{tabular}%
}
        \caption{Memory usage per batch for each architecture}
        \label{tab:memoryusage}
\end{table}

\subsection{Implementation}

We used Python to implement the models and evaluate their performance. Initially we implemented all models and training by hand using the popular NumPy library \cite{oliphant2006guide}, to fully understand the possible extensions to make to tree LSTMs. On a larger scale, the training of the models in our NumPy implementation became prohibitively expensive, and as such we re-implemented all code again in the popular deep learning library PyTorch \cite{paszke2017automatic}. Pytorch allowed us to do the mathematical computations on GPUs easily, thus speeding up model training. Furthermore, Pytorch's underlying neural network implementation uses dynamic knowledge graphs, which permitted us to implement a variable size tree LSTM architecture.
All code used will be published at \href{https://github.com/CedricCook/}{https://github.com/CedricCook}.

\section{Experiments}

Because of the enormous memory usage that we observe, we decide to only run all 4 variants of the models on a small dataset, whereas we can run the first three models on a large dataset. Both datasets retain the same distribution of merchants across countries, and both datasets include only one web page per merchant to guarantee a diverse dataset. The large dataset contains a total of 30 484 datapoints, and the small dataset contains 1512 datapoints. The train/validation/test splits for both datasets remain at 64\%/20\%/16\%.

\subsection{Comparison of classification on large dataset}

We will train the Fully Connected network, the Monodirectional bottom-up tree LSTM and the Bidirectional tree LSTM using top-town features, all on the same large training dataset. All models are as described above including loss function and optimizer. All models will be trained for 150 epochs.

\subsection{Comparison of classification on small dataset}

In order to also assess the performance of the 4th model, our novel bidirectional Tree LSTM using bottom-up embeddings, we will train all four models on the small training dataset as well. All 4 models are trained for 150 epochs using the same optimizer and loss function.

\subsection{Model ablation}

In order to quantify the contribution of well engineered features to the models' performances, we perform feature ablation. The original set of features that makes up the input vector of each node potentially depends strongly on the features that are easily distinguishable, such as the bounding box of the element. The bounding box is encoded in our feature vector as (top\_left\_x, top\_left\_y, width, height). In this experiment, we remove these 4 scalars from the feature vector of all nodes, and train 3 models again entirely, each on the small dataset. Finally we evaluate the small test set, also with the bounding boxes removed, on the trained models.

\section{Results \& Discussion}

For each experiment, we trained all models relevant to that experiment for 150 epochs, and evaluated the model on the validation data set at the end of each epoch. We store the state of the model, of the epoch where the Categorical Cross Entropy loss on the validation set was lowest.
Finally, we evaluated this saved model state on the test set relevant to each experiment. This evaluation provides us with the model's achieved F1-score per target class. Below follow the interpretations of the results of these evaluations.

\subsection{Comparison of classification on large dataset}

\begin{table}[b!]
\centering
\resizebox{\textwidth}{!}{%
\begin{tabular}{|l||r|r|r|}
\hline
\rowcolor[HTML]{EFEFEF} 
{\color[HTML]{333333} } & {\color[HTML]{333333} Fully Connected} & {\color[HTML]{333333} Monodirectional} & {\color[HTML]{333333} Bidirectional - Features} \\ \hline
\rowcolor[HTML]{FFFFFF} 
negative                & \textbf{0.8827}                        & 0.8595                                 & 0.8823                                          \\ \hline
\rowcolor[HTML]{FFFFFF} 
name                    & 0.8345                                 & 0.8208                                 & \textbf{0.8594}                                 \\ \hline
\rowcolor[HTML]{FFFFFF} 
cart                    & 0.9404                                 & 0.9548                                 & \textbf{0.9615}                                 \\ \hline
\rowcolor[HTML]{FFFFFF} 
price                   & \textbf{0.8408}                        & 0.8304                                 & 0.8369                                          \\ \hline
\rowcolor[HTML]{FFFFFF} 
addtocart               & \textbf{0.8677}                        & 0.8159                                 & 0.848                                           \\ \hline
\rowcolor[HTML]{FFFFFF} 
mainpicture             & 0.9357                                 & \textbf{0.9488}                        & 0.9461                                          \\ \hline
\rowcolor[HTML]{FFFFFF} 
subjectnode             & 0.9681                                 & \textbf{0.9766}                        & 0.9656                                          \\ \hline\hline
\rowcolor[HTML]{FFFFFF} 
micro avg               & 0.8897                                 & 0.8784                                 & \textbf{0.8941}                                 \\ \hline
\rowcolor[HTML]{FFFFFF} 
macro avg               & 0.8957                                 & 0.8867                                 & \textbf{0.9}                                    \\ \hline
\rowcolor[HTML]{FFFFFF} 
weighted avg            & 0.8902                                 & 0.8792                                 & \textbf{0.8943}                                 \\ \hline
\end{tabular}%
}
\caption{F1-scores per class on the web page element classification task, as evaluated on the large test set. The three models were trained on the large training set.}
\label{tab:resultsLargeDataset}
\end{table}

The results of the three models performance evaluation on web page element classification task after training on the large dataset are shown in Table \ref{tab:resultsLargeDataset}. The first observation that we can make is that the three models all achieve high per-class F1 scores. 
If we look specifically at the scores per target class we can interpret where the performance differences are. For example, when predicting an element to the \textsl{name} target, we observe that the Bidirectional model performs much better than the other two models. 
The name of a product on a web page presents itself in a very heterogeneous fashion; it can often be any tag type such as \texttt{div}, \texttt{p}, \texttt{h1} or any other \texttt{h} element. Other features such as the coordinates of the \textsl{name} element may also vary widely across web pages. However, the bidirectional model performs significantly better on this target, suggesting that the information it captured from the root-subtree path is a helpful distinguishing factor.

On the contrary, elements such as \textsl{price} and \textsl{addtocart} are very often in the same region of a product page, notably the bottom right.
This would indicated that their bounding box coordinates are in some cluster, and thus without further context information, the Fully Connected model can learn to recognise these elements efficiently. Furthermore, we see that the \textsl{subjectnode} is learnt best by the Monodirectional model. The \textsl{subjectnode}, on average, is much higher up in the tree than other elements such as the \textsl{price}, and thus the root-subtree path is shorter. This may be an issue for Bidirectional - Features model that relies on that top-down information. However, on average the Bidirectional - Features model performs reasonably well, and thus its average F1-scores are higher than both other models.

As we also want to evaluate our novel Bidirectional - Embeddings tree LSTM model, we performed the same task and training, but instead on the small dataset. Not only do we gain some interesting insights about the novel model, but we also will interpret the difference in performance of the models we have already seen when presented with much less data for training.

\subsection{Comparison of classification on small dataset}
The results of the three models performance evaluation on web page element classification task after training on the small dataset are shown in Table \ref{tab:resultsSmallDataset}. We include once again the same three models as before, Fully Connected, Monodirectional and Bidirectional - Features, but we also add a column for the results of our novel Bidirectional - Embeddings tree LSTM. NB, these scores are not the same as table \ref{tab:resultsLargeDataset}, instead all models were completely retrained on the small dataset. Our first observation is that the first three models perform worse on nearly every target class, compared to when trained on the large data. Section \ref{sec:dataQuantity} goes into further analysis of the impact of the quantity of training data used.

\begin{table}[b]
\centering
\resizebox{\textwidth}{!}{%
\begin{tabular}{|l||r|r|r|r|}
\hline
\rowcolor[HTML]{EFEFEF} 
             & Fully Connected & Monodirectional & Bidirectional - Features & Bidirectional - Embeddings \\ \hline
\rowcolor[HTML]{FFFFFF} 
negative     & 0.6538          & 0.6944          & \textbf{0.7848}          & 0.7324                     \\ \hline
\rowcolor[HTML]{FFFFFF} 
name         & 0.6486          & \textbf{0.6761} & 0.6176                   & 0.6667                     \\ \hline
\rowcolor[HTML]{FFFFFF} 
cart         & 0.9062          & \textbf{0.9231}          & 0.8923                   & 0.9091            \\ \hline
\rowcolor[HTML]{FFFFFF} 
price        & 0.7458          & 0.6765          & \textbf{0.7606}          & 0.7273                     \\ \hline
\rowcolor[HTML]{FFFFFF} 
addtocart    & \textbf{0.6301} & 0.5789          & 0.5915                   & 0.6                        \\ \hline
\rowcolor[HTML]{FFFFFF} 
mainpicture  & 0.9492          & 0.9455          & 0.9455                   & \textbf{0.9643}            \\ \hline
\rowcolor[HTML]{FFFFFF} 
subjectnode  & 0.9804          & 0.9455          & 0.9811                   & \textbf{0.9811}            \\ \hline\hline
\rowcolor[HTML]{FFFFFF} 
micro avg    & 0.7778          & 0.7619          & \textbf{0.7835}          & 0.7817                     \\ \hline
\rowcolor[HTML]{FFFFFF} 
macro avg    & 0.7877          & 0.7771          & 0.7962                   & \textbf{0.7973}            \\ \hline
\rowcolor[HTML]{FFFFFF} 
weighted avg & 0.7788          & 0.7637          & \textbf{0.7874}          & 0.7852                     \\ \hline
\end{tabular}%
}
\caption{F1-scores per class on the web page element classification task, as evaluated on the small test set. The four models were trained on the small train set.}
\label{tab:resultsSmallDataset}
\end{table}

As before, the best results are distributed across the various models. If we observe the scores of the Bidirectional - Embeddings model closely, we notice that even though they are not the highest for all target classes, the model almost systematically achieves second best score and never ranks lowest. All three earlier models perform very similar on the prediction of the \textsl{mainpicture}, but are all outperformed by the novel method. The \textsl{mainpicture} is another element that has relatively generic node features, making it hard to classify. It seems that the high amount of context propagated from around the tree to this element helps the Bidirectional - Embeddings model to achieve a higher score here. Overall, the novel model exhibits respectable, yet not stellar, performance when trained on such a small dataset. However, the overall performance is close to the maxima shown by the other models. This lead us to investigate how the novel model may perform when trained on a greater dataset.

\subsection{Impact of data quantity on model performance}
\label{sec:dataQuantity}

To accurately interpret the difference in performance of models when trained on more or less data, we present the delta of the scores achieved per class by the models, when trained on the large and small dataset. We do this for the Fully Connected, Monodirectional and Bidirectional - Features models as they were all trained on both datasets. We present the deltas of the two evaluations in table \ref{tab:datasetSizeDiff}. The main observation we make from these results is that except for the \textsl{negative} class, the two latter models have a larger classification performance improvement than the Fully Connected model when being trained on a larger dataset. For example, in the cases of the \textsl{name}, \textsl{cart} and \textsl{mainpicture}, one of the latter two models has a performance improvement of almost double with respect to the much simpler Fully Connected model. The Monodirectional and Bidirectional - Features models are much more complex in their quantity of parameters. We theorize that the simpler Fully Connected model learns fast on a small dataset, but reaches a lower performance celing. It seems that the more complex models take longer to learn but can eventually reach a higher maximum performance on more data. This leads us to believe that this may be the case as well for the Bidirectional - Embeddings model. This novel model is by far the most complex of all four models discussed here, and it is plausible that it may outperform the other three methods when presented with enough training data.

\begin{table}[b]
\centering
\resizebox{\textwidth}{!}{%
\begin{tabular}{|l||r|r|r|}
\hline
\rowcolor[HTML]{EFEFEF} 
            & Fully Connected & Monodirectional & Bidirectional - Features \\ \hline
\rowcolor[HTML]{FFFFFF} 
negative    & \textbf{0.2289} & 0.1651          & 0.0975                   \\ \hline
\rowcolor[HTML]{FFFFFF} 
name        & 0.1859          & 0.1447          & \textbf{0.2418}          \\ \hline
\rowcolor[HTML]{FFFFFF} 
cart        & 0.0342          & 0.0317          & \textbf{0.0692}          \\ \hline
\rowcolor[HTML]{FFFFFF} 
price       & 0.095           & \textbf{0.1539} & 0.0763                   \\ \hline
\rowcolor[HTML]{FFFFFF} 
addtocart   & 0.2376          & 0.237           & \textbf{0.2565}          \\ \hline
\rowcolor[HTML]{FFFFFF} 
mainpicture & -0.0135         & \textbf{0.0033} & 0.0006                   \\ \hline
\rowcolor[HTML]{FFFFFF} 
subjectnode & -0.0123         & \textbf{0.0311} & -0.0155                  \\ \hline
\end{tabular}%
}
\caption{Increase in F1-score per class for the three models, when trained on the large dataset vs. when trained on the small dataset.}
\label{tab:datasetSizeDiff}
\end{table}

\subsection{Model ablation}
In our final experiment, we attempt to quantify the impact of the local input features to the overall performance of the models. To do so, we have removed all features considering the bounding box of an element from all elements, thus leaving 66 out of the total of 70 features. We now train the Fully Connected, Monodirectional, and Bidirectional - Features models with these fewer features, on the small training dataset. The results from their evaluations on the small test set are shown in Table \ref{tab:resultsAblation}. We notice that almost all scores for all combinations of targets and models are significantly lower. From this we deduce that the models rely heavily on the quality of the features local to each node. Another observation is that the Fully Connected layer suffers the most, because the local features form a much bigger proportion of the information the network can decide on, compared to the tree LSTM models. The Bidirectional - Features model performs the best specifically on the \textsl{cart} and \textsl{price} targets. This may well be attributable to the depth of these two elements, as they are on average the deepest in the tree of all the classes we consider. Therefore, the Bidirectional - Features model has a long root to subtree path to compute on, which provides more information than the small subtrees of these elements that the Monodirectional model infers on.
The overall take away from this section is not that the bounding box is an important component to web page element classification, but rather that feature engineering is a very important problem to address in this space, and that tree LSTM models perform better than non-structural models on DOM tree data.

\begin{table}[h]
\centering
\resizebox{\textwidth}{!}{%
\begin{tabular}{|l||r|r|r|}
\hline
\rowcolor[HTML]{EFEFEF} 
             & Fully Connected & Monodirectional & Bidirectional - Features \\ \hline
\rowcolor[HTML]{FFFFFF} 
negative     & 0.0             & \textbf{0.5528} & 0.2609                   \\ \hline
\rowcolor[HTML]{FFFFFF} 
name         & 0.2798          & \textbf{0.4074} & 0.3273                   \\ \hline
\rowcolor[HTML]{FFFFFF} 
cart         & 0.0             & 0.4673          & \textbf{0.5283}          \\ \hline
\rowcolor[HTML]{FFFFFF} 
price        & 0.0             & 0.4407          & \textbf{0.4776}          \\ \hline
\rowcolor[HTML]{FFFFFF} 
addtocart    & \textbf{0.3143} & 0.0             & 0.2469                   \\ \hline
\rowcolor[HTML]{FFFFFF} 
mainpicture  & 0.0             & 0.0             & 0.0                      \\ \hline
\rowcolor[HTML]{FFFFFF} 
subjectnode  & 0.8235          & \textbf{0.8966} & 0.8667                   \\ \hline\hline
\rowcolor[HTML]{FFFFFF} 
micro avg    & 0.2646          & \textbf{0.4719} & 0.4279                   \\ \hline
\rowcolor[HTML]{FFFFFF} 
macro avg    & 0.2025          & 0.395           & \textbf{0.3868}          \\ \hline
\rowcolor[HTML]{FFFFFF} 
weighted avg & 0.1825          & \textbf{0.3973} & 0.3752                   \\ \hline
\end{tabular}%
}
\caption{F1-scores per class on the web page element classification task, as evaluated on the small test set. The four models were trained on an ablated training set, where the features corresponding to the element bounding box were removed.}
\label{tab:resultsAblation}
\end{table}

\subsection{Runtime analysis}

To conclude the results section, we discuss the run times of the 4 different models on the small and large datasets. We present the time it took each model on each dataset to train for 150 epochs in Table \ref{tab:runtime}. 

\begin{table}[h!]
\centering
\resizebox{\textwidth}{!}{%
\begin{tabular}{|l||r|r|r|}
\hline
\rowcolor[HTML]{EFEFEF} 
                           & Small Dataset - 150 epochs & Large Dataset - 150 epochs & Times Slower \\ \hline
\rowcolor[HTML]{FFFFFF} 
Fully Connected            & 42                         & 852                        & 20.8         \\ \hline
\rowcolor[HTML]{FFFFFF} 
Monodirectional            & 12058                      & 273059                     & 22.6         \\ \hline
\rowcolor[HTML]{FFFFFF} 
Bidirectional - Features   & 13738                      & 293655                     & 21.4         \\ \hline
\rowcolor[HTML]{FFFFFF} 
Bidirectional - Embeddings & 516323                     & N/A                        & N/A          \\ \hline
\end{tabular}%
}
\caption{Training time for all models on the small and large datasets, in seconds. The \textsl{Times Slower} column is the training time on the large dataset divided by the training time on the small dataset.}
\label{tab:runtime}
\end{table}

For the three models that were trained on both datasets, the increase in training time from the small dataset to the large dataset is close to linear with the increase in quantity of data, as the large dataset contains exactly 20 times more data than the small set. From these results we can only estimate that the same will hold true for the Bidirectional - Embeddings method, and thus it would take 10 326 400 seconds, or just under 120 days to train this model for 150 epochs on the large dataset, unless any further optimisations are made. 

\subsubsection{Results over time}
\begin{figure}[h]
    \centering
    \includegraphics[width=0.8\linewidth]{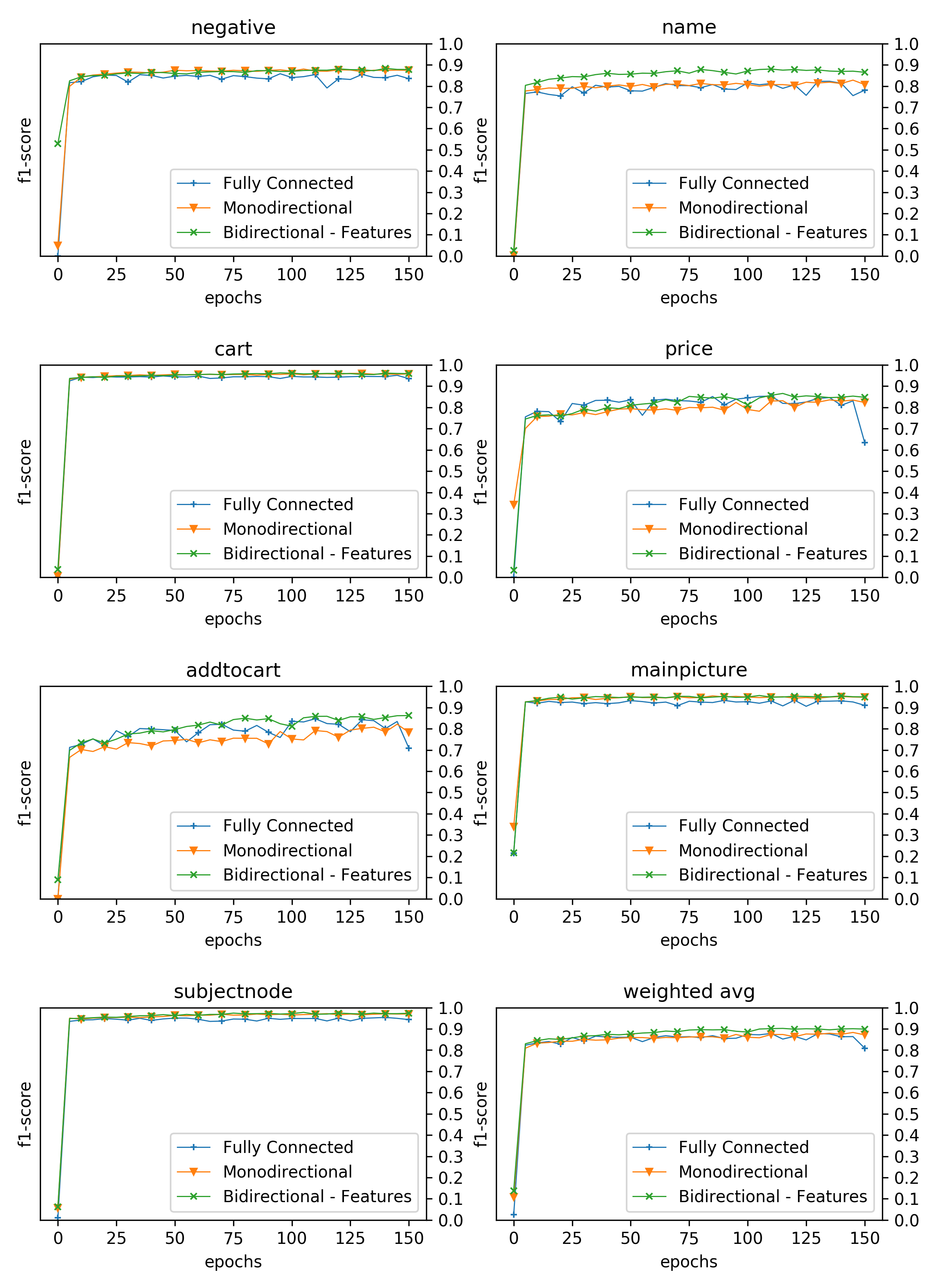}
    \caption{F1-scores per class as evaluated on the validation set during training. These models were trained on the large dataset.}
    \label{fig:validationScoresLarge}
\end{figure}

\begin{figure}[h]
    \centering
    \includegraphics[width=1\linewidth]{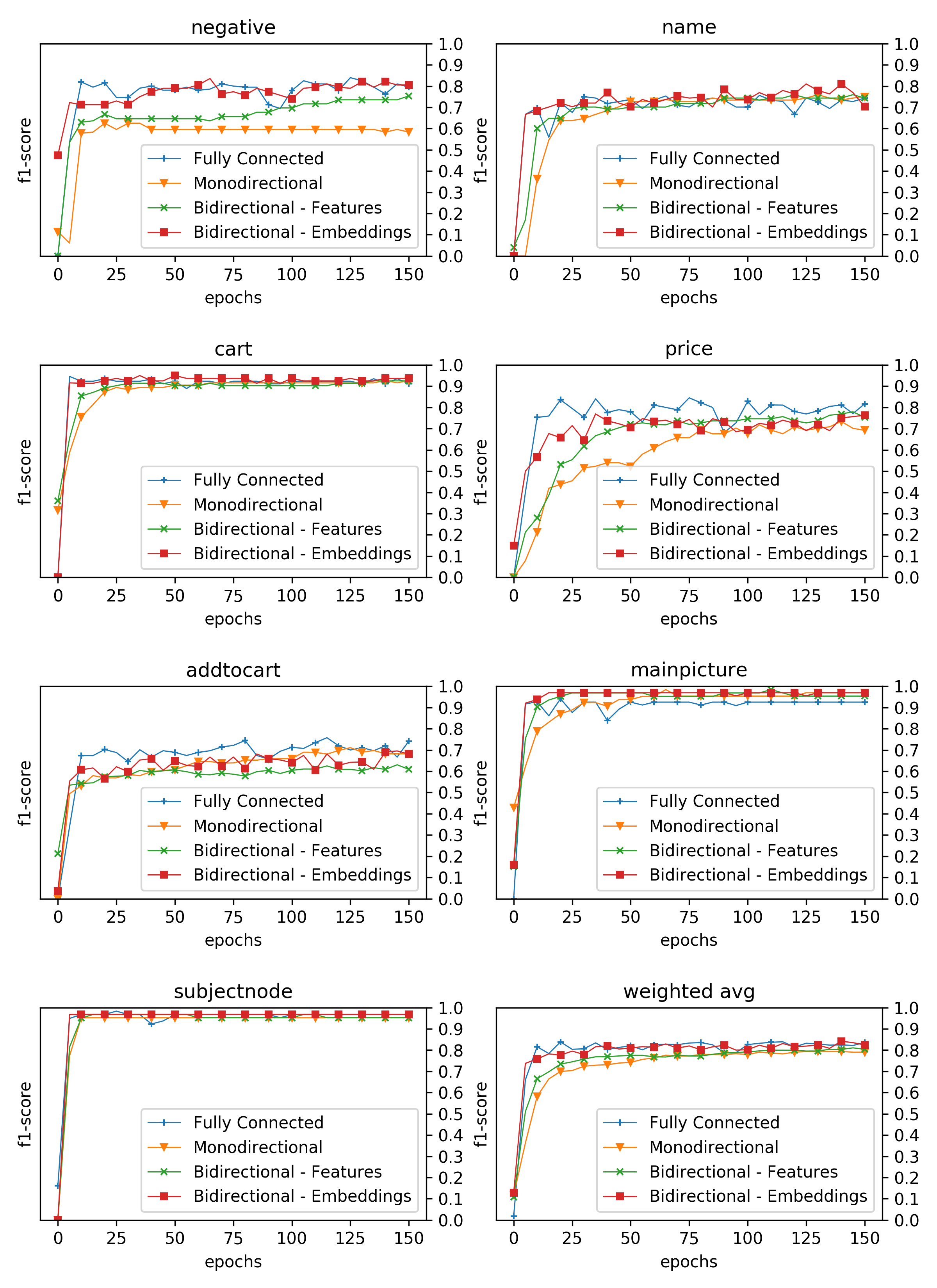}
    \caption{F1-scores per class as evaluated on the validation set during training. These models were trained on the small dataset.}
    \label{fig:validationScoresSmall}
\end{figure}

\begin{figure}[h]
    \centering
    \includegraphics[width=1\linewidth]{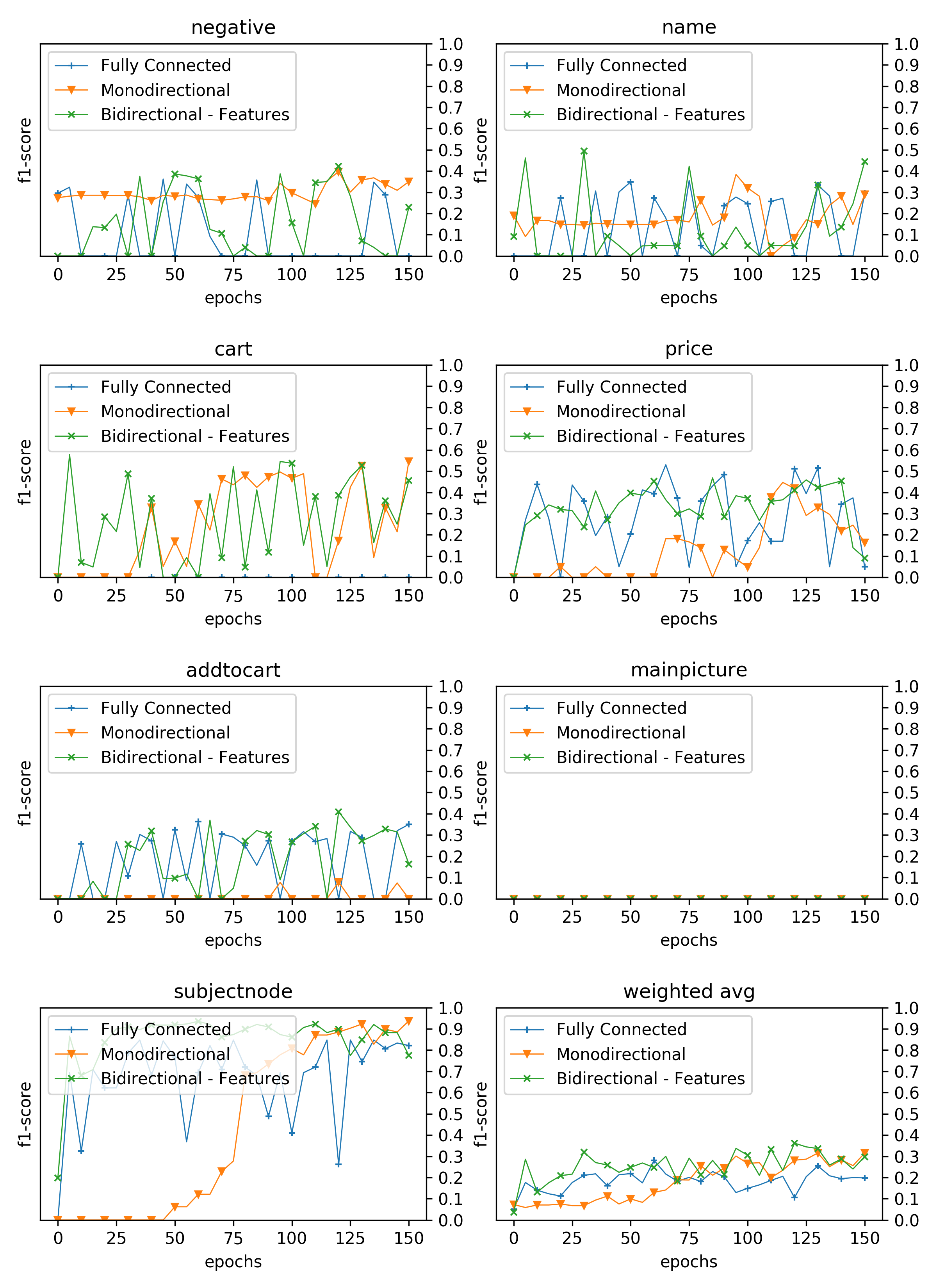}
    \caption{F1-scores per class as evaluated on the validation set during training. These models were trained on an ablated version of the small dataset, where the input features did not include the boundingbox of the elements.}
    \label{fig:validationScoresAblation}
\end{figure}

In the appendix we also present the plots that indicate each model's performance as evaluated on the validation set at the end of each epoch in Figures \ref{fig:validationScoresLarge}, \ref{fig:validationScoresSmall}, \ref{fig:validationScoresAblation}. This data gives us an insight to whether a model has stopped learning or may achieve a better performance when provided with more time.

\section{Future Work}

From both the qualitative and quantitative results of our experiments, a few future work items appear that may be addressed.

\subsection{Implementation Optimisation}
To address the exorbitant training time manifested by the Bidirectional - Embeddings tree LSTM model, optimisations to the implementation of the model and its training must be made. Many of such optimisations lie in the processing of the data as it is directed through the neural network computation graph. One major improvement to the training time may be the use of multiple GPUs in parallel, where we only used one single GPU for the computations.

\subsection{Feature Engineering}

The main interest of this research was to transport a somewhat known method from NLP and apply it to web element classification space. Because of this focus, feature engineering was almost entirely left out of scope. It has become clear through the results of the experiments presented in this research that the features are of very high importance to the performance of the model.
Some features of the DOM tree that we did not consider but that are low hanging fruit in terms of increasing model performance are text and image content. Regardless of the final task of the model, a subtree is most likely represented better when the embedding can make use of the actual content of each node. Future use of this model may extend the current choice of feature vector by either using some preprocessing or pretrained models to generate features for the text and image content, or prepend layers to the model that train embeddings for text with a bidirectional LSTM and image embeddings with a Convolutional Network. These prepended layers can be trained jointly with the rest of the models.

\subsection{Bidirectional method improvement}

\subsubsection{Dedicated bottom-up tree - LSTM unit}
In our proposed novel bidirectional method we use one kernel to generate bottom-up embeddings that are used in two distinct ways. We use this kernel to make bottom-up embeddings that are trained to represent the subtree and the elements contained in it, which is analog to training the deep contents of the subtree. Secondly, we use this same kernel to generate embeddings in the same fashion for the whole tree, which we subsequently use in our top-down method in order to take context from the rest of the tree to the subtree. The result of this configuration is that the kernel of the bottom-up component is essentially being trained for two different uses, which may hurt model performance. Therefore, we hypothesise that an improvement may be found in using two bottom-up tree LSTMs in parallel. The first remains unchanged to as it was proposed earlier in this report, and only computes on the subtree of interest. The second tree LSTM cell will use a completely separate kernel. This secondary cell will compute on the entire tree, and generate embeddings that may be used by the top-down component. This new configuration should separate the two tasks and allow both kernels to converge better.

\subsubsection{Independent context embeddings}

Both in the current bidirectional embedding model and the version proposed in the previous paragraph, the bottom-up component is computed over the entire tree, including the subtree of interest. We hypothesize that, especially for larger subtrees, the context may contribute more effectively if it is only context independent from the the subtree itself. One approach to achieve independent context is to make a copy of the tree structure from which the subtree originates, and remove the subtree from that copy. Then, a separated bottom-up tree LSTM cell is computed over the entire tree in the copy, and the top-down component is computed over the embeddings generated on this copy. The final embedding of the subtree is now a concatenation of the bottom-up representation of the subtree, and a comprehensive representation of the independent context of the subtree.

\section{Conclusion}
In this Master Thesis we introduced a generalized method for learning efficient context-aware representations of subtrees. The results from our experiments have shown that tree LSTM models can be applied successfully to structured web data, and we demonstrated that such models outperform conventional deep learning methods for the application of web element classification. We introduced a novel extension to tree LSTM that captures the context of a subtree and generates a more efficient representation of a subtree, and showed its preliminary performance on the web as well as potential it has when optimized. Such context-aware representations may be generalized and used for further applications such as state estimators in reinforcement learning on the web.

\bibliographystyle{apalike}
\bibliography{report}

\newpage
\section{Appendix}

\subsection{Syntactic parse trees}
\label{apx:parsetrees}

SemEval (Semantic Evaluation) is an ongoing series of evaluations of computational semantic analysis systems. The evaluations are intended to explore the nature of meaning in language. While meaning is intuitive to humans, transferring those intuitions to computational analysis has proved elusive.
Distributional Semantic Models (DSMs) approximate the meaning of words with vectors summarizing their co-occurrence patterns in text corpora. Compositional DSMs (CDSMs) are an extension to DSMs, with a purpose of representing meaning of phrases by composing the distributional representation of the words the sentences contain. Semantics of sentences in human language is shown not be contained only in linear sequences of words, but also in the parse tree of the phrases.
These parse trees can be of two types, Dependency Parse Trees or Constituency Parse Trees, where dependency trees have a variable number of children per node, and each node contains a word, whereas constituency trees only contain words on the leaf nodes, and are n-ary. The non-leaf nodes in constituency trees have grammatical syntax meaning in the phrases.

\subsection{Tree LSTM Recursion Overview}

\subsection{Performance over time during training}

The following plots show the F1-scores of each model as evaluated at the end of each training epoch. One subplot indicated the F1-score of a model of a specific target class.

\end{document}